\pdfoutput=1

\documentclass[11pt]{article}

\usepackage[final]{coling}
\usepackage{tabularx}
\usepackage{comment}
\usepackage{xcolor}
\usepackage{booktabs}
\usepackage{tcolorbox}
\usepackage{float} 
\def\red#1{\textcolor{red}{#1}}

\usepackage{times}
\usepackage{latexsym}
\usepackage{url}

\usepackage[T1]{fontenc}

\usepackage[utf8]{inputenc}
\usepackage{multirow}
\usepackage{caption}
\usepackage{subcaption}
\usepackage{mwe}

\usepackage{microtype}

\usepackage{inconsolata}

\usepackage{graphicx}

\usepackage{booktabs} 

\usepackage{amsmath} 
\title{GEAR: A Simple \textsc{Generate, Embed, Average and Rank} Approach for Unsupervised Reverse Dictionary}

\author{
\textbf{
Fatemah Almeman\textsuperscript{1,2}},
 \textbf{ Luis Espinosa-Anke\textsuperscript{1,3}}
\\
\\
\textsuperscript{1}CardiffNLP, School of Computer Science and Informatics, Cardiff University, UK \\
\textsuperscript{2}College of Computer and Information Sciences, Princess Nourah bint Abdulrahman University, KSA \\
\textsuperscript{3}AMPLYFI, UK \\
 \small{
  \textbf{Correspondence:} \href{mailto:email@domain}{almemanf@cardiff.ac.uk}
 }
}

\begin{document}
\maketitle
\begin{abstract}
Reverse Dictionary (RD) is the task of obtaining the most relevant word or set of words given a textual description or dictionary definition. Effective RD methods have applications in accessibility, translation or writing support systems. Moreover, in NLP research we find RD to be used to benchmark text encoders at various granularities, as it often requires word, definition and sentence embeddings. In this paper, we propose a simple approach to RD that leverages LLMs in combination with embedding models. Despite its simplicity, this approach outperforms supervised baselines in well studied RD datasets, while also showing less over-fitting. We also conduct a number of experiments on different dictionaries and analyze how different styles, registers and target audiences impact the quality of RD systems. We conclude that, on average, untuned embeddings alone fare way below an LLM-only baseline (although they are competitive in highly technical dictionaries), but are crucial for boosting performance in combined methods.\footnote{The code and data are available at \url{https://github.com/F-Almeman/GEAR_RD}} 
\end{abstract}

\section{Introduction}
\label{sec:introduction}

Reverse dictionary (RD), conceptual dictionary or concept lookup is the task to returning a word or set of suitable words given a text description or definition \cite{zock2004word,zock2004wordb}. This NLP task broadly equates to the psycholinguistic notion of ``lexical access'' and ``tip of the tongue problem'' \cite{brown1966tip}, and is crucial for better understanding of the  mental lexicon, i.e., how we as humans store meaning and render semantic representations into words \cite{wanner1996lexical,zock2010deliberate}. In practical terms, RD systems are helpful assistants to writers and translators \cite{yan-etal-2020-bert}, and can have significant impact on the linguistic experience of language learners with limited vocabulary \cite{zhang2020multi} or people affected with anomic aphasia \cite{benson1979neurologic}. From an NLP perspective, RD serves as a tool for fine-tuning and evaluating text encoders \cite{hill2016learning,pilehvar2019importance,zhang2020multi,chen2022unified} and, recently, as a probe for understanding Large Language Models' (LLMs) internal representations \cite{xu2024tip}. Despite its usefulness, research in RD is currently limited in two fronts. First, RD benchmarks are mostly sourced from WordNet \cite{miller1995wordnet} and the Oxford Dictionary, and little is known about the effectiveness of RD methods on other resources or languages - with a few notable exceptions such as the multilingual experiments in \citet{yan-etal-2020-bert}. This is problematic because the generalization ability of models optimized for these two standard and over-utilized resources might not reflect modern, acquired, rare, evolving or technical terminologies. And second, because there is a surprising lack of work exploiting the generative capabilities of LLMs to improve over embedding-only baselines. While \citet{tian2024prompt} propose to leverage LLMs for RD using prompt engineering and obtain good results, their approach requires fine-tuning a text generation model in the first stage, and the final set of predicted terms may not correspond to the vocabulary of the dictionary in question, making this approach hard to apply on large-scale real-world resources\footnote{At the time of writing this manuscript, the English Wiktionary has over 7.5M entries (with over 30M entries across all languages), making embedding search a prerequisite on any realistic RD method. \url{https://en.wikipedia.org/wiki/Wiktionary}}.

In this paper, we make the following contributions. First, we propose \textsc{GEAR} (\textbf{g}enerate, \textbf{e}mbed, \textbf{a}verage and \textbf{r}ank), a novel lightweight and unsupervised method for RD that utilizes an LLM for generating a set of candidates given an input definition, and pools their corresponding embeddings into a vector used for KNN search. This fast and highly scalable method outperforms heavily tuned supervised baselines (including those leveraging LLMs), setting a new state of the art in two (the two that require generalization) out of three test sets in Hill's dataset \cite{hill2016learning}. We also evaluate \textsc{GEAR}, alongside other unsupervised baselines, on a diverse set of dictionaries, from WordNet to Urban Dictionary\footnote{\url{https://www.urbandictionary.com/}}, and perform an in-depth analysis of the strengths and weaknesses of these methods as per the domain, register and target audience of each dictionary.

\section{Related Work}
\label{sec:relatedwork}

This paper is relevant to computational semantics, and more concretely, on the interaction between dictionaries and NLP. Therefore, we cover, first, prominent work in this intersection (Section \ref{sec:rw:dictionaries}),  and second, we give an account of RD works (Section \ref{sec:rw:reversedictionary}).

\subsection{Dictionaries and NLP}
\label{sec:rw:dictionaries}

Dictionaries and NLP have a healthy relationship. Indeed, dictionaries have proven to be suitable resources for improving text processing pipelines at different stages. For example, for improving Word Sense Disambiguation (WSD) systems based on BERT \cite{devlin2018bert} by fine-tuning them on \textit{context-definition} pairs \cite{huang2019glossbert}, or by combining into one loss function different pre-training strategies, e.g., discriminating between correct and wrong definitions via contrastive learning, or dictionary entry prediction \cite{yu2022dict,chen2022dictbert}. Beyond BERT, \citet{be:20} fine-tuned BART \cite{lew:19} on example-definition pairs, and reported high results in intrinsic benchmarks and, more importantly, used their DM system for downstream NLP, specifically WSD as well as word-in-context \cite{pilehvar-camacho-collados-2019-wic} classification. This strategy was further adopted to add an interpretability layer to semantic change detection via definition generation \cite{giulianelli2023interpretable}. Other examples of successfully marrying dictionaries and NLP systems include training a unified vector space of words, definitions and ``mentions in context'' \cite{gajbhiye2024amended}, which were successfully used for ontology completion; or fine-tuning an LLM such as LLAMA2 \cite{touvron2023llama} on WordNet's semantic relations and definitions, which then can be flexibly used in several lexical semantics tasks \cite{moskvoretskii2024taxollama}.



\subsection{Reverse Dictionary}
\label{sec:rw:reversedictionary}

Concerning RD, this is a task with a long tradition in lexical semantics, with early methods exploiting hand-crafted rules \cite{bi:04,shaw2011building} for extracting textual features. It was \citet{hill2016learning} who introduced RNNs as suitable architectures that complemented bag of words representations, as well as a dataset specific to RD sourced from different resources, such as WordNet, Webster's Dictionary and Wiktionary, among others. From here, the usage of neural networks first, and more specific, pre-trained transformer encoders like BERT later, have dominated the RD landscape. Among the former, let us highlight, e.g., \citet{pilehvar2019importance}, who integrates WordNet senses and supersenses as an additional signal, improving over text-based embeddings alone. Further, \citet{zhang2020multi} propose a multi-channel model comprising a sentence encoder based on BiLSTMs and multiple linguistically motivated predictors such as word category (using WordNet's taxonomy), morpheme or sememe prediction, whereas \citet{chen2021towards} directly replaces word embeddings with synset embeddings, optionally leveraging examples of usage. An immediate limitation of the above works is their reliance on a sense inventory such as WordNet, which has proven to work very well for modeling in-domain terminologies, less so for enabling generalization (cf. Table \ref{tab:hill_comparison}).

Other works have exploited multiple but related tasks in the broad ``embedding a dictionary'' paradigm, e.g., by combining definition generation and RD with reconstruction tasks via autoencoders \cite{chen2022unified}, or have fine-tuned T5 \cite{raffel2020exploring} with excellent results \cite{mane2022wordalchemy}. More recently, LLMs have unexpectedly been introduced into RD. For example, in a two-stage approach where a fine-tuned LLM first generates a set of candidates which are then passed in a subsequent prompt to a generator for outputting the final set of predictions \cite{tian2024prompt}. Finally, from a more ``probing'' perspective, RD has been used to gain insights into LLMs representations via conceptual inference, showing that they encode information about object categories as well as fine-grained features \cite{xu2024tip}.

As we can see from the above survey, dictionaries play an important role in NLP today, and RD is a suitable probe for pre-training and evaluating text encoders as well as LLMs. However, as we mentioned in Section \ref{sec:introduction}, there has been little work in expanding existing benchmarks beyond Hill's dataset, especially in terms of different domains and registers. Moreover, no works have explored the seemingly simple generate-then-embed approach, so that many of the practical drawbacks of LLMs (hallucinations and context length limitations, to name a few) could be alleviated, but still using their ability to generate suitable candidate embeddings for KNN search.

\section{The \textsc{GEAR} Method}
\label{sec:gear}

In this section, we give a brief description of \textsc{GEAR}, a novel, \textit{very} simple, lightweight, and, more importantly, unsupervised method for RD. We denote any dictionary as $D = \{(d_i, T_i) | i = 1, \ldots, N\}$, where $d_i$ is a definition, $T_i = \{t_{i1}, t_{i2}, \ldots, t_{ik_i}\}$ is the set of corresponding terms\footnote{It is important to account for a one-to-many relationship at this point because we will be reporting experiments on combinations of multiple resources.} (i.e., entries in the dictionary), and $k_i \geq 1$ the number of terms associated with $d_i$. From here, \textsc{GEAR} consists on four simple steps. First, \textbf{g}enerate, where, given an input definition $d_i$, an (LLM) generates a set of possible terms $G = \{g_1, g_2, \ldots, g_m\}$. All throughout this paper, we use \textsc{gpt-4o-mini} \cite{achiam2023gpt}\footnote{\url{https://openai.com/index/gpt-4o-mini-advancing-cost-efficient-intelligence/}.}, which we prompt in three different ways  to thoroughly explore any differences in performance that might arise from deeper descriptions of the task. The full details for these prompts can be found in Appendix \ref{sec:appendix:prompts}), however, at a high level, they can be summarized as follows:

\begin{itemize}
    \item Base prompt 1 (\textbf{bp1}): includes a short description of the resource: such as \textit{Given the definition \{definition\}, generate a ranked list of \{k\} terms, with the first term being the most related to the definition, assuming they are from the \{resource\} dictionary. \{resource\} is \{resource description\}.}
    
    \item Base prompt 2 (\textbf{bp2}): the same as bp1 but it also includes a sample of terms and definitions from the specified resource to help the model understand the type of terms it should generate: such as \textit{These are some examples of definitions and terms in this dictionary: \{examples\}.}

    \item Reasoning prompt (\textbf{rp}): this final prompt extends bp2 by requesting the generation of examples alongside the terms to explore whether having the LLM 'reason' before answering leads to improved results. Specifically, this part is added:  \textit{For each term, provide an example usage in a sentence that matches the style and scope of \{dictionary\}.}
\end{itemize}

In the next \textbf{e}mbed step, a text encoder $f: \mathcal{V} \rightarrow R^{n}$ maps each term in $G$ to a vector representation in an $n$-dimensional space. We use SBERT \cite{reimers2019sentence} and the Instructor model \cite{su2022one} to obtain term embeddings and evaluate their performance for comparison (see Section \ref{sec:results}). The resulting matrix $\mathbf{E}_G = [\mathbf{e}_1, \mathbf{e}_2, \ldots, \mathbf{e}_m]^\top \in R^{m \times n}$, where $\mathbf{e}_i = f(g_i)$, is then mean pooled (or \textbf{a}veraged) as follows:  $\mathbf{\bar{e}} = \frac{1}{m} \sum_{i=1}^m \mathbf{e}_i$. Finally, in the \textbf{r}ank step, given $T = \bigcup_{i=1}^N T_i$, which denotes the set of all unique terms in $D$, we perform KNN search via cosine similarity over $T$ with  $\mathbf{\bar{e}}$. The performance of \textsc{GEAR}, just like any other search-based approach, can be evaluated using Information Retrieval metrics that account for different scenarios, e.g., the rank of the first correct term with Mean Reciprocal Rank (MRR), or the proportion of correct terms at different cutoffs with Precision at k (P@k). 

\begin{table*}[!t]
\scriptsize
\resizebox{\textwidth}{!}{%
\begin{tabular}{p{6cm} p{4cm} p{2cm}}
\toprule
\multicolumn{1}{c}{\textbf{Definition}} & \multicolumn{1}{c}{\textbf{Terms}} & \multicolumn{1}{c}{\textbf{Sources}} \\ \midrule
Alert and fully informed                     
& [knowing, knowledgeable]
& [WN, WN]                         
\\\midrule

River in singapore
& [Geylang River, Singapore River]   
& [WP, WP]       
\\ \midrule

In the middle of the week
& [midweek] 
& [Wik, CHA]
\\ \midrule

A type of shotgun
& [12 gauge, greener]
& [Urban, CHA]                           
\\ \midrule
An arsonist
& [arsonite, torchman, incendiary]
& [Wik, Urban, Mul]
\\ \midrule

Any supply that is running low
& [low supply, short supply]
& [Hei++, Hei++]
\\ \midrule

A term used to describe something so awesome the only way it could be better is if it was between two slices of bread
& [ass kicking sandwitch]
& [Urban]
\\ \midrule

An abnormal accumulation of air in the pleural space (the space between the lungs and the chest cavity) that can result in the partial or complete collapse of a lung
& [Primary Spontaneous Pneumothorax]
& [Sci]                    
\\ \bottomrule
\end{tabular}%
}
\caption{\texttt{3D-EX} examples. Note that one definition could map to more than one term, which in turn can come from different dictionaries. Also note the range of styles and domains. (WordNet (WN), Wikipedia (WP), Wiktionary (Wik), MultiRD (Mul), Sci-definition (Sci))}
\label{tab:examples}
\end{table*}

\section{Data}
\label{sec:data}

In this paper we are concerned not only with exploring the usefulness of \textsc{GEAR} when compared to existing baselines, we are also interested in developing an understanding of what kind of lexical resource poses greater challenges to this method as well as its components alone. For this reason, we perform experiments on two distinct but complementary datasets.

As a first evaluation set, we use the three test sets from the \citealt{hill2016learning} dataset (Section \ref{sec:exp:hill}) to compare \textsc{GEAR} with other published RD methods: the seen set, which includes 500 word-definition pairs from the training set to evaluate recall; the unseen set, containing 500 pairs where both the words and definitions are excluded from training; and the description set, consisting of 200 words with with human-written descriptions. Both the \textit{unseen} and \textit{human description} datasets are suitable for determining the generalization ability of any tested method. Secondly, we report performance on the different dictionaries included in \texttt{3D-EX} \cite{al:23} (Section \ref{sec:exp:3dex}), a comprehensive resource that integrates multiple dictionaries and organizes them into \texttt{<word,definition>} pairs and \texttt{<word,definition,example>} triplets\footnote{While dictionary examples are a valuable resource for improving text representations, we leave them out of our experiments in order to limit the number of components to test.}. We convert this dataset into a suitable RD format, namely \texttt{<definition, list of terms>} and perform two types of splits: a \textit{random split} and a \textit{source split}. In the random split, the data is split randomly into 60\% for training, 20\% for validation, and 20\% for testing. In the source split, all definitions from each source in the dataset are extracted into separate datasets, and then each dataset is split randomly into training, validation, and test sets using the same 60\%, 20\%, and 20\% ratio. Despite the unsupervised nature of our work, we still conduct all our experiments on the test splits alone to enable comparison in further iterations with supervised methods. Table \ref{tab:examples} shows examples of entries in \texttt{3D-EX}  in its new RD format, and illustrates the diversity in register, domain and style within the resource. For the benefit of the reader, we provide a brief description of each resource integrating \texttt{3D-EX} below:

\begin{table*}[!t]
\renewcommand{\arraystretch}{1.5}  
\large
\resizebox{\textwidth}{!}{%
\begin{tabular}{l|>{\centering}p{1cm}>{\centering}p{2.5cm}>
{\centering\arraybackslash}p{1cm}|>{\centering}p{1cm}>{\centering}p{2.5cm}>{\centering\arraybackslash}p{1cm}|>{\centering}p{1cm}>{\centering}p{2.5cm}>{\centering\arraybackslash}p{1cm}}
\toprule
\multirow{2}{*}{\textbf{\large Model/Method}} & \multicolumn{3}{c|}{\textbf{\large Seen Definition}} & \multicolumn{3}{c|}{\textbf{\large Unseen Definition}} & \multicolumn{3}{c}{\textbf{\large Description}} \\ \cline{2-10} 
 & \textbf{mr} & \textbf{acc@k} & \textbf{rv} 
 & \textbf{mr} & \textbf{acc@k} & \textbf{rv} 
 & \textbf{mr} & \textbf{acc@k} & \textbf{rv} \\ \midrule
OneLook       & \multicolumn{1}{r|}{\textbf{0}}  & \multicolumn{1}{r|}{66/.94/.95} & \multicolumn{1}{r|}{200} 
              & \multicolumn{1}{r|}{-} & \multicolumn{1}{r|}{-} & \multicolumn{1}{r|}{-} & 
              \multicolumn{1}{r|}{5.5} & \multicolumn{1}{r|}{.33/.54/.76} & \multicolumn{1}{r}{332} \\ \hline
BOW           & \multicolumn{1}{r|}{172} & \multicolumn{1}{r|}{.03/.16/.43} & \multicolumn{1}{r|}{414}
              & \multicolumn{1}{r|}{248} & \multicolumn{1}{r|}{.03/.13/.39} & \multicolumn{1}{r|}{424} 
              & \multicolumn{1}{r|}{22}  & \multicolumn{1}{r|}{.13/.41/.69} & \multicolumn{1}{r}{308} \\
             
RNN           & \multicolumn{1}{r|}{134} & \multicolumn{1}{r|}{.03/.16/.44} & \multicolumn{1}{r|}{375}
              & \multicolumn{1}{r|}{171} & \multicolumn{1}{r|}{.03/.15/.42} & \multicolumn{1}{r|}{404}
              & \multicolumn{1}{r|}{17} & \multicolumn{1}{r|}{.14/.40/.73} & \multicolumn{1}{r}{274} \\
              
RDWECI        & \multicolumn{1}{r|}{121} & \multicolumn{1}{r|}{.06/.20/.44} & \multicolumn{1}{r|}{420}
              & \multicolumn{1}{r|}{170} & \multicolumn{1}{r|}{.05/.19/.43} & \multicolumn{1}{r|}{420}
              & \multicolumn{1}{r|}{16} & \multicolumn{1}{r|}{.14/.41/.74} & \multicolumn{1}{r}{306} \\
              
SuperSense    & \multicolumn{1}{r|}{378} & \multicolumn{1}{r|}{.03/.15/.36} & \multicolumn{1}{r|}{462}
              & \multicolumn{1}{r|}{465} & \multicolumn{1}{r|}{.02/.11/.31} & \multicolumn{1}{r|}{454} 
              & \multicolumn{1}{r|}{115} & \multicolumn{1}{r|}{.03/.15/.47} & \multicolumn{1}{r}{396} \\
              
MS-LSTM       & \multicolumn{1}{r|}{\textbf{0}} & \multicolumn{1}{r|}{\textbf{.92/.98/.99}} & \multicolumn{1}{r|}{\textbf{65}}
              & \multicolumn{1}{r|}{276} & \multicolumn{1}{r|}{.03/.14/.37} & \multicolumn{1}{r|}{426} 
              & \multicolumn{1}{r|}{1000} & \multicolumn{1}{r|}{.01/.04/.18} & \multicolumn{1}{r}{404} \\ 
Multi-channel & \multicolumn{1}{r|}{16}  & \multicolumn{1}{r|}{.20/.44/.71} & \multicolumn{1}{r|}{310}  
              & \multicolumn{1}{r|}{54}  & \multicolumn{1}{r|}{.09/.29/.58} & \multicolumn{1}{r|}{358}   
              & \multicolumn{1}{r|}{2}   & \multicolumn{1}{r|}{.32/.64/.88} & \multicolumn{1}{r}{203} \\ 
BERT          & \multicolumn{1}{r|}{\textbf{0}} & \multicolumn{1}{r|}{.57/.86/.92} & \multicolumn{1}{r|}{240} 
              & \multicolumn{1}{r|}{18} & \multicolumn{1}{r|}{.20/.46/.64} & \multicolumn{1}{r|}{418} 
              & \multicolumn{1}{r|}{1}  & \multicolumn{1}{r|}{.36/.77/.94} & \multicolumn{1}{r}{94} \\
               
RoBERTa       & \multicolumn{1}{r|}{\textbf{0}} & \multicolumn{1}{r|}{.57/.84/.92} & \multicolumn{1}{r|}{228} 
              & \multicolumn{1}{r|}{37} & \multicolumn{1}{r|}{.10/.36/.60} & \multicolumn{1}{r|}{405} 
              & \multicolumn{1}{r|}{1}  & \multicolumn{1}{r|}{.43/.85/.96} & \multicolumn{1}{r}{46} \\ \midrule \midrule
GEAR\_bp1     & \multicolumn{1}{r|}{\textbf{0}} & \multicolumn{1}{r|}{.66/.84/.96} & \multicolumn{1}{r|}{200.122}               
              & \multicolumn{1}{r|}{\textbf{0}} & \multicolumn{1}{r|}{\textbf{.70/.88/.97}} & \multicolumn{1}{r|}{\textbf{180.955}}  
              & \multicolumn{1}{r|}{\textbf{0}} & \multicolumn{1}{r|}{.89/.99/.99} & \multicolumn{1}{r}{70.5334} \\
GEAR\_bp2     & \multicolumn{1}{r|}{\textbf{0}} & \multicolumn{1}{r|}{.71/.88/.97} & \multicolumn{1}{r|}{170.451}                    
              & \multicolumn{1}{r|}{\textbf{0}} & \multicolumn{1}{r|}{.65/.82/.95} & \multicolumn{1}{r|}{225.324}                                 
              & \multicolumn{1}{r|}{\textbf{0}} & \multicolumn{1}{r|}{\textbf{.93/.99/1}} & \multicolumn{1}{r}{\textbf{1.57314}} \\ 
GEAR\_rp      & \multicolumn{1}{r|}{\textbf{0}} & \multicolumn{1}{r|}{.70/.87/.96} & \multicolumn{1}{r|}{185.97}      
              & \multicolumn{1}{r|}{\textbf{0}} & \multicolumn{1}{r|}{.66/.86/.96} & \multicolumn{1}{r|}{190.8}               
              & \multicolumn{1}{r|}{\textbf{0}} & \multicolumn{1}{r|}{.91/.99/1}   & \multicolumn{1}{r}{1.7837} \\ \hline
\end{tabular}%
}
\caption{\textsc{GEAR} results on the Hill's dataset compared to competitor models (using the Instructor model for embeddings, as it achieves the best performance), according to median rank (mr), accuracy@k (acc @ 1/100/1000), and rank variance (rv). Baselines results are from \citet{zhang2020multi} and \citet{yan-etal-2020-bert}.}
\label{tab:hill_comparison}
\end{table*}

\begin{itemize}

    \item \textbf{WordNet}: A lexical database that groups words into synsets (synonym sets) with definitions, lemmas, examples, and word relations \cite{miller1995wordnet, fe:13}. WordNet is commonly used as a sense inventory in NLP \cite{ag:07, zh:22}. 

    \item \textbf{CHA}: A dataset containing words, definitions, and examples from the Oxford Dictionary \cite{ch:19}. It has been used for dictionary modeling \cite{be:20} and to evaluate the quality of WordNet's examples \cite{al:22}.
    
    \item \textbf{Wikipedia}: A free online encyclopedia collaboratively created by contributors around the world \cite{ya:16}.
    
    \item \textbf{Wiktionary}: A web-based dictionary that provides detailed information on words, including definitions, examples, and pronunciation \cite{ba:22}.
    
    \item \textbf{Urban}: A crowd-sourced platform that focuses on slang and informal language not usually covered in traditional dictionaries \cite{st:20}. 
    
    \item \textbf{CODWOE}: The English dataset from the CODWOE (Comparing Dictionaries and Word Embeddings) SemEval 2022 shared task \cite{mi:22}.
    
    \item \textbf{Sci-definition}: A dataset specifically created to produce definitions for scientific terms with different levels of complexity \cite{august-etal-2022-generating}.

    \item \textbf{Webster's Unabridged}: A version of Webster's dictionary \cite{no:90} available through Project Gutenberg \cite{va:09}, offering definitions and additional notes for English words.

    \item \textbf{MultiRD}: A dataset developed by \cite{le:19} to evaluate a reverse dictionary model, using the English dictionary data from \citet{hill2016learning}.

    \item \textbf{Hei++}: A dataset created by \citeauthor{be:20} that pairs human-made definitions with adjective-noun phrases, based on the test split of the HeiPLAS dataset \cite{ha:15}.
\end{itemize}

\section{Main Experiments}
\label{sec:exp}

\subsection{\textsc{GEAR} on Hill's dataset}
\label{sec:exp:hill}

We introduce two sets of experiments on Hill's dataset. The first one uses an LLM alone to perform RD by directly generating terms from input definitions. The second experiment integrates the LLM with embedding models to form \textsc{GEAR}, which enhances performance across the board. For evaluation, we compare \textsc{GEAR} results with the following baselines: (1) OneLook, which is the most popular commercial RD system \cite{zhang2020multi}; (2) BOW and RNN with rank loss \cite{hill2016learning}, which are neural models where BOW uses a bag-of-words approach and RNN employs Long Short-Term Memory (LSTM); (3) RDWECI \cite{mo:18}, which improves BOW by adding category inference; (4) SuperSense \cite{pilehvar2019importance}, which advances BOW by using pre-trained sense embeddings; (5) MS-LSTM \cite{kartsaklis-etal-2018-mapping}, which enhances RNN with WordNet synset embeddings and a multi-sense LSTM; (6) Multi-channel \cite{zhang2020multi}; and (7) BERT and RoBERTa \cite{yan-etal-2020-bert}, which are trained to generate the target word for the RD task. We use three evaluation metrics based on previous work: median rank of target words (lower is better), accuracy of target words in the top K results (higher is better), and rank variance (lower is better). Table \ref{tab:hill_comparison} shows that \textsc{GEAR} outperforms all baselines on both the unseen definition set and the description set. We can also observe that MS-LSTM performs effectively on the seen definition set but not on the description set, showing its limited ability to generalize \cite{zhang2020multi}.

\subsection{\textsc{GEAR} on 3D-Ex}
\label{sec:exp:3dex}
Despite the new SoTa established on Hill's dataset, we are also interested in exploring core components such as LLMs or embeddings on other resources. To this end, and taking \texttt{3D-Ex} as a test bed, we apply the same set of experiments as described in Section \ref{sec:exp:hill} and introduce a new experiment that uses only different text embeddings (without the \textbf{g}enerate step). This will serve as a baseline to what extent \textsc{GEAR} can provide an improvement over these basic and untuned approaches.

In terms of experimental setup, unless otherwise specified, we always evaluate a fixed-length ranked list of k terms for a given definition, which are then compared to the gold terms. As for evaluation metrics, we use two different metrics:

\begin{itemize}
    \item \textit{Mean Reciprocal Rank} (MRR), which measures the position of the first correct result in a list of outcomes, is defined as \begin{equation}
    \mbox{{MRR}} = \frac{1}{|Q|}\sum_{i=1}^{|Q|}\frac{1}{rank_i}
    \end{equation}

    \noindent where $Q$ is a sample of experiment runs and $rank_i$ refers to the rank position of the \textit{first} relevant outcome for the \textit{i}th run. 

    \item \textit{Precision @ k} (P@k), which calculates the precision of relevant items within the top kk positions of a ranked list, is defined as follows:
    \begin{equation}
    \mbox{{Precision@k}} = \frac{1}{k} \sum_{i=1}^{k} \text{rel}_i
    \end{equation}

    \noindent where \(\text{rel}_i\) is 1 if the item at position \(i\) is relevant and 0 otherwise. We now introduce the different embedding models we use in our experiments.

    \end{itemize}

\paragraph{\textbf{Embeddings}} We evaluate the performance of different embedding models, which we select considering factors such as adoption among the community, performance in open benchmarks such as MTEB \cite{muennighoff2023mteb}, availability in the HuggingFace\footnote{\url{https://huggingface.co/}} hub, as well as being of manageable size. These models are:

\begin{itemize}
    \item SBERT \cite{reimers2019sentence} models, namely {\small{\texttt{all-MiniLM-L6-v2}}}, {\small{\texttt{all-distilroberta-v1}}} and {\small{\texttt{all-mpnet-base-v2}}}.

    \item Jina Embeddings \cite{ji:23} which is a language model that has been trained using Jina AI's Linnaeus-Clean dataset that contains query-document pairs. We use {\small{\texttt{jina-embedding-b-en-v1}}} and {\small{\texttt{jina-embedding-l-en-v1}}}.

    \item General Text Embeddings (GTE) model \cite{li:23} which is trained on a large-scale corpus of relevance text pairs from different domains. In this work we use {\small{\texttt{gte-large}}}.

    \item Instructor \cite{su2022one}. It generates text embeddings for different tasks (such as classification or retrieval) and domains (such as science or finance) based on task instructions. We use {\small{\texttt{instructor-large}}}. We examined three different variants of instructions for encoding terms and definitions: (1) no instructions provided; (2) using a general description for the target text (i.e., ``Represent the sentence:'' and ``Represent the word:''); and (3) applying dictionary-specific instructions for the target texts (i.e., ``Represent the dictionary definition:'' and ``Represent the sentence: the dictionary entry:'').

    \item Universal AnglE Embedding \cite{an:23}, another instruction-based encoder, and which we use with the same configurations as Instructor. We use {\small{\texttt{UAE-Large-V1}}}.
   
\end{itemize}

\section{Results and Analysis}
\label{sec:results}

\subsection{Hill's dataset}
\label{sec:hill_results}

Table \ref{tab:GEAR_results_hill} shows how the performance improves with the \textsc{GEAR} method. In the first part of the table, where candidates are evaluated without any embeddings, the likelihood of having the target term in the top 5 candidates is low. Among the models tested with \textsc{GEAR}, the Instructor model, which encodes terms as dictionary entries based on instructions, performs best, which gains of above 10\% on the seen and unseen splits, but negligible differences on the human descriptions (presumably because these already get a good sentence embedding from sentence BERT, on one hand, and also because they might not be accurately described as dictionary resources, which is what we used as an instruction for Instructor). For prompt effectiveness, we found that adding the requirement to generate a dictionary example, and despite its usefulness in other settings, does not improve over base prompt 2, which simply provides as input a few exemplars. 

\begin{table}[ht]
\centering
\small
\begin{tabular}{llllcc}
\toprule
\textbf{Model} & \textbf{Split} & \textbf{Prompt} & \textbf{ACC@1} & \textbf{ACC@5} \\
\midrule
\multirow{9}{*}{-} &
\multirow{3}{*}{S.} &
bp1 & 26.8 & 44.0 \\
& & bp2 & 30.4 & 50.2 \\
& & rp  & 29.2 & 46.8 \\ 
\cmidrule{2-5}
& \multirow{3}{*}{U.} &
bp1 & 30.0 & 47.2 \\
& & bp2 & 33.4 & 53.8 \\
& & rp  & 33.4 & 49.8 \\
\cmidrule{2-5}
& \multirow{3}{*}{D.} &
bp1 & 70.0 & 77.0 \\
& & bp2 & 72.5 & 83.0 \\
& & rp  & 72.0 & 81.5 \\
\midrule
\multirow{9}{*}{SBERT} &
\multirow{3}{*}{S.} &
bp1 & 57.8 & 68.2 \\
& & bp2 & 62.4 & 73.0 \\
& & rp  & 60.6 & 72.2 \\ 
\cmidrule{2-5}
& \multirow{3}{*}{U.} &
bp1 & 59.0 & 70.8 \\
& & bp2 & 63.8 & 77.0 \\
& & rp  & 61.6 & 75.2 \\
\cmidrule{2-5}
& \multirow{3}{*}{D.} &
bp1 & 90.5 & 96.5 \\
& & bp2 & 93.5 & 97.5 \\
& & rp  & 94.0 & 98.0 \\
\midrule
\multirow{9}{*}{Instructor} &
\multirow{3}{*}{S.} &
bp1 & 66.0 & 80.6 \\
& & bp2 & 71.4 & 84.6 \\
& & rp  & 70.4 & 84.0 \\ 
\cmidrule{2-5}
& \multirow{3}{*}{U.} &
bp1 & 64.6 & 79.0 \\
& & bp2 & 70.0 & 83.4 \\
& & rp  & 66.4 & 82.4 \\
\cmidrule{2-5}
& \multirow{3}{*}{D.} &
bp1 & 89.5 & 98.0 \\
& & bp2 & 92.5 & 98.5 \\
& & rp  & 91.5 & 99.0 \\
\bottomrule
\end{tabular}
\caption{Performance comparison of LLMs (no embeddings models, top block) and \textsc{GEAR} methods (middle and bottom block) across various prompts in Hill's dataset. S.: Seen split, U.: Unseen split, and D.: human description split. Prompt types are bp1 (Base Prompt 1), bp2 (Base Prompt 2), and rp (Reasoning Prompt).}
\label{tab:GEAR_results_hill}
\end{table}



\begin{table}[!ht]
\centering
\small  
\resizebox{\columnwidth}{!}{%
\begin{tabular}{l lcccc}
\toprule
\textbf{Model} & \textbf{Prompt} & \textbf{MRR} & \textbf{P@1} & \textbf{P@3} & \textbf{P@5} \\
\midrule
\multirow{3}{*}{-}& 
bp1   & 28.27 &	24.58 & 10.64 & 6.91 \\ &
bp2   & 30.21 &26.18 & 11.39& 7.42 \\ &
rp    &  30.99 & 26.98 & 11.70 & 75.89 \\ 
\hline
\multirow{3}{*}{SBERT} &
bp1  & 40.61 & 33.75 & 17.36 & 11.96    \\
& bp2  &  43.01 & 36.21 & 18.32 & 12.59 \\
& rp   & 44.21 & 37.09 & 18.92 & 12.99 \\ 
\hline
\multirow{3}{*}{Instructor} & 
bp1  & 43.47 & 36.41 & 18.00 & 12.24 \\
& bp2 & 45.58 & 38.67 & 18.76 & 12.73   \\
& rp  &  46.37	& 39.31 & 19.09 & 12.98  \\
\bottomrule
\end{tabular}%
}
\caption{Performance comparison of LLMs (no embeddings models) and \textsc{GEAR} methods across different models and prompts in \texttt{3D-EX}, showing the average score across different dictionaries.}
\label{tab:GEAR_results_3dex}
\end{table}

\subsection{\texttt{3D-EX} Dataset}
Table \ref{tab:GEAR_results_3dex} presents the average MRR, P@1, P@3, and P@5 for each prompt across different resources in \texttt{3D-EX}, comparing two methods: one without embedding models and the other using the \textsc{GEAR} method. As previously demonstrated in Hill's results (Section \ref{sec:hill_results}), the \textsc{GEAR} method, particularly with the Instructor model, achieves the best performance, showing improvements in MRR and P@1 ranging from 3 to 6 points. Concerning the type of prompt, interestingly, we found the sophistication of the prompt to matter the most when combined with embedding models (where we see an improvement of around 7\% MRR from base to reasoning), but only 2\% when prompting alone is considered. 

\begin{figure*}[!ht]
    \centering
    \begin{tabular}{cc}
        \begin{tabular}{c}
            \fbox{\resizebox{0.4\textwidth}{!}{\includegraphics{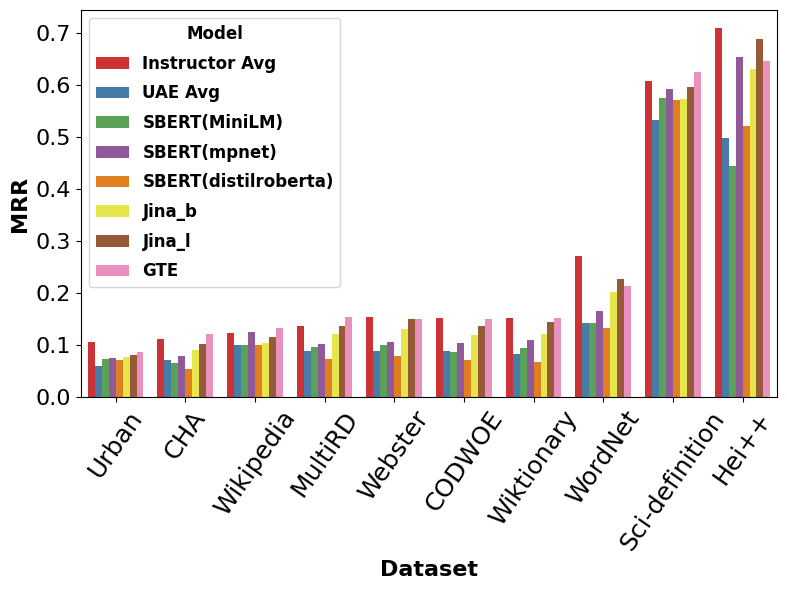}}}\\
            \small (a) MRR
        \end{tabular}
        &
        \begin{tabular}{c}
            \fbox{\resizebox{0.4\textwidth}{!}{\includegraphics{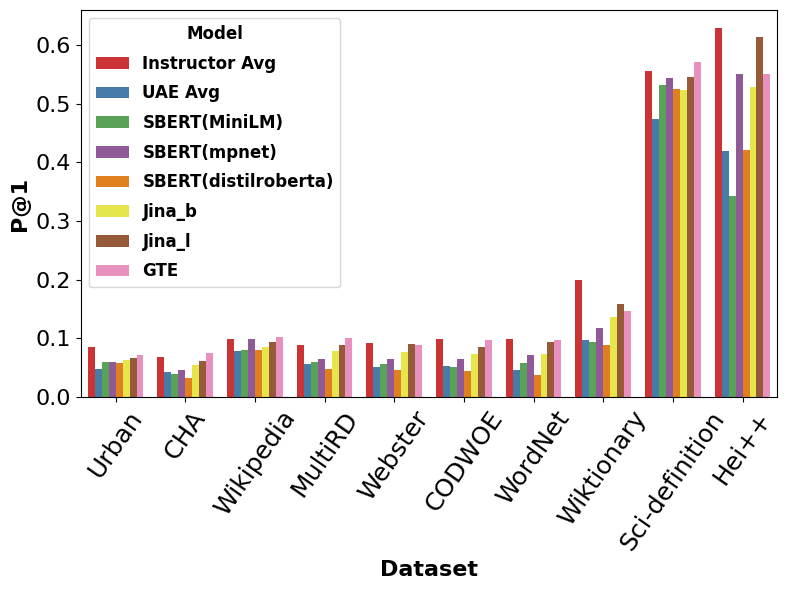}}}\\
            \small (b) P@1
        \end{tabular}
        \\[1ex]
        \begin{tabular}{c}
            \fbox{\resizebox{0.4\textwidth}{!}{\includegraphics{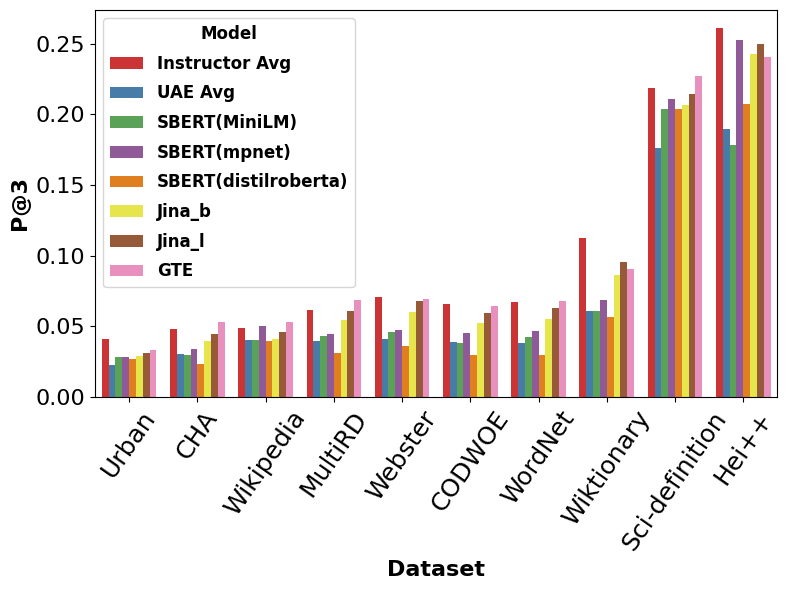}}}\\
            \small (c) P@3
        \end{tabular}
        &
        \begin{tabular}{c}
            \fbox{\resizebox{0.4\textwidth}{!}{\includegraphics{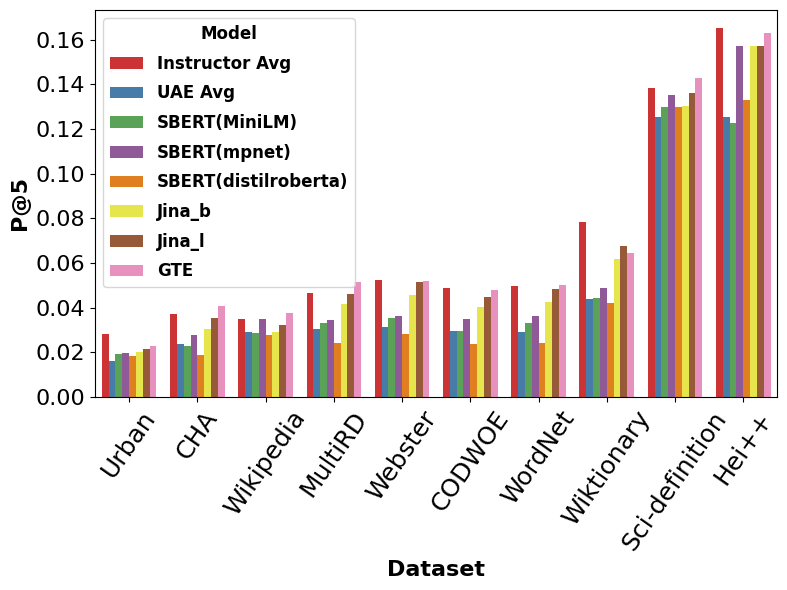}}}\\
            \small (d) P@5
        \end{tabular}
    \end{tabular}
    \caption{Performance comparison for various embedding models across different metrics in \texttt{3D-EX}}
    \label{fig:3dex}
\end{figure*}

Table \ref{tab:embed_results}
shows the average MRR, P@1, P@3, and P@5 for each embedding model mentioned in Section \ref{sec:exp:3dex}. Results are much lower compared to those achieved with the \textsc{GEAR} method, as well as below prompting alone. In Figure \ref{fig:3dex}, we illustrate the performance across these different datasets, and verify that Hei++ and Sci-definition datasets have higher values, while Urban and CHA show lower values. This variation is likely due to the nature of the entries in Hei++ and Sci-definition, designed to capture more specialized and unique terms. We see, interestingly, that while Instructor embeddings alone are consistently outperforming the rest, they particularly shine in WordNet, which suggest that WordNet embeddings may benefit from additional context to the encoder, since it has been shown that WordNet's definitions and examples are perhaps too short to be informative \cite{almeman2022putting,giulianelli2023interpretable}.

\begin{table}[H]
\centering
\small  
\begin{tabular}{lrrrr}
\toprule
\textbf{Model} & \textbf{MRR} & \textbf{P@1} & \textbf{P@3} & \textbf{P@5} \\
\midrule
Instructor (dict. - dict.)   & 24.88  & 19.80  & 9.82   & 6.73  \\
Instructor (gen. - dict.)    & 24.81  & 19.40  & 9.96   & 6.78  \\
Instructor (gen. - no)       & 24.73  & 19.72  & 9.78   & 6.68  \\
Instructor (dict. - no)      & 24.51  & 19.55  & 9.68   & 6.60  \\
Instructor (gen. - gen.)     & 24.32  & 19.17  & 9.68   & 6.65  \\
Instructor (dict. - gen.)    & 24.15  & 19.05  & 9.55   & 6.61  \\
Instructor (no - no)         & 24.03  & 19.23  & 9.45   & 6.47  \\
Instructor (no - dict.)      & 23.92  & 18.91  & 9.51   & 6.50  \\
Jina (large)                 & 23.78  & 18.96  & 9.33   & 6.40  \\
GTE (large)                  & 23.47  & 18.18  & 9.38   & 6.56  \\
Instructor (no - gen.)       & 23.24  & 18.31  & 9.20   & 6.37  \\
UAE (gen. - gen.)            & 22.93  & 17.63  & 9.25   & 6.41  \\
UAE (dict. - gen.)           & 21.99  & 16.96  & 8.77   & 6.16  \\
UAE (gen. - dict.)           & 21.79  & 16.80  & 8.71   & 6.12  \\
UAE (dict. - dict.)          & 20.90  & 15.87  & 8.45   & 5.94  \\
Jina (base)                  & 20.85  & 16.11  & 8.38   & 5.81  \\
all-mpnet-base-v2            & 20.14  & 15.96  & 7.96   & 5.45  \\
UAE (gen. - no)              & 18.91  & 14.48  & 7.66   & 5.36  \\
UAE (dict. - no)             & 17.77  & 13.55  & 7.20   & 5.01  \\
all-MiniLM-L6-v2             & 17.04  & 13.07  & 6.84   & 4.82  \\
all-distilroberta-v1         & 16.55  & 13.05  & 6.54   & 4.50  \\
UAE (no - gen.)              & 8.98   & 7.25   & 3.51   & 2.39  \\
UAE (no - dict.)             & 8.10   & 6.50   & 3.16   & 2.15  \\
UAE (no - no)                & 5.34   & 3.81   & 2.24   & 1.58  \\
\bottomrule
\end{tabular}

\caption{Comparing different embedding models without any support from an LLM-based generation step, showing the average score across \texttt{3D-EX} dictionaries.}
\label{tab:embed_results}
\end{table}

\section{Analyzing \textsc{GEAR} Components}

\paragraph{Generation with Open Source LLM} To ensure that our approach is both effective across different models and accessible for future research, we repeated the \textsc{GEAR} experiment on the Hill's dataset using Llama \cite{touvron2023llama}, specifically \texttt{Llama 3.1-70B} \footnote{\url{https://huggingface.co/meta-llama/Llama-3.1-70B}}, an open-source, freely available, and highly capable language model. As demonstrated in Table \ref{tab:llama}, our method continues to outperform the competitor systems in 2 out of 3 datasets.

\begin{table}[H]
\centering
\small
\begin{tabular}{llllcc}
\toprule
\textbf{Split} & \textbf{Prompt} & \textbf{ACC@1} & \textbf{ACC@10} & \textbf{ACC@100}\\
\midrule

\multirow{3}{*}{S.} &
bp1 & 55.6	& 67.4 & 76.8 \\
& bp2 & 69.8 & 83.6	& 94.2 \\
& rp  & 76.2 & 89 & 97 \\ 
\cmidrule{1-5}
\multirow{3}{*}{U.} &
bp1 & 68.8 & 84.4 & 92.4 \\
& bp2 & 67.2 & 85 & 93.2 \\
& rp  & 69.8 & 87.8	& 95.2 \\
\cmidrule{1-5}
\multirow{3}{*}{D.} &
bp1 & 88 & 97.5	& 99.5 \\
& bp2 & 88.5 & 98.5	& 99.5 \\
& rp  & 91.5 & 100 & 100 \\

\bottomrule
\end{tabular}
\caption{\textsc{GEAR} results using Llama for candidates generation and Instructor model for embeddings across different prompts in Hill's dataset. S.: Seen split, U.: Unseen split, and D.: human description split. Prompt types are bp1 (Base Prompt 1), bp2 (Base Prompt 2), and rp (Reasoning Prompt).}
\label{tab:llama}
\end{table}

\begin{figure*}[t]
    \centering
    \begin{tabular}{c c}
        \begin{tabular}{c}
            \includegraphics[width=0.35\textwidth]{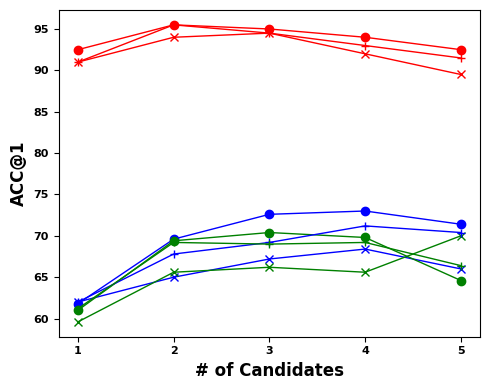}\\
            \small (a) Accuracy@1
        \end{tabular}
        &
        \begin{tabular}{c}
            \includegraphics[width=0.35\textwidth]{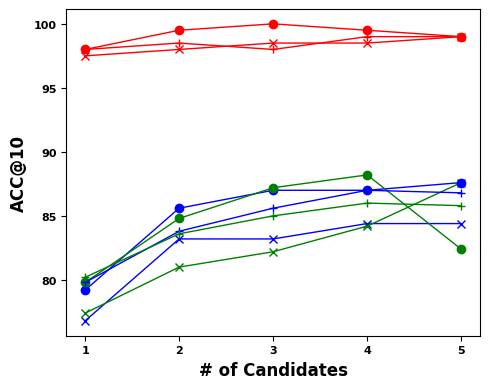}\\
            \small (b) Accuracy@10
        \end{tabular}
        \\[1ex] 
        
        \multicolumn{2}{c}{ 
            \begin{tabular}{c}
                \includegraphics[width=0.5\textwidth]{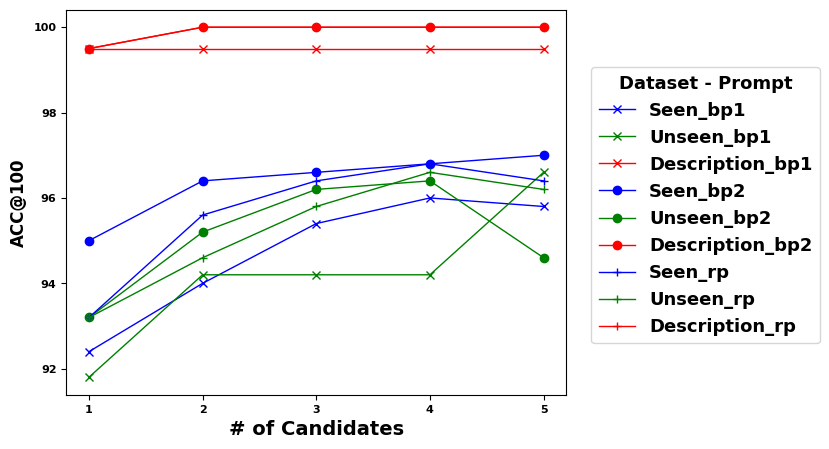}\\
                \small (c) Accuracy@100
            \end{tabular}
        }
    \end{tabular}
    \caption{A comparison of the performance on Hill's splits, evaluating the number of candidates in the \textbf{g}enerate step, which are then subsequently averaged to produce the input vector for KNN search.}
    \label{fig:num_can}
\end{figure*}

\paragraph{Different Pooling Methods} In order to gain a deeper understanding of the effect of the number of candidates produced during the generation step of our \textsc{GEAR} method, we plot the performance metrics across different candidate values for all three splits and for precision @ k. Figure \ref{fig:num_can} shows that using just one candidate is not optimal, while averaging over 2 or 3 candidates provides better results, sometimes outperforming all the 5. These results also suggest that while we could have tuned the candidate number on a development set, with the tools we tested (\texttt{gpt-4o-mini}; and Instructor and SBERT), it seems proven that performance plateaus at only a handful of generated terms. 

Additionally, We explored the effectiveness of max pooling instead of averaging the generated term embeddings. These experiments did not provide any improvements over the averaging method results shown in Table \ref{tab:hill_comparison}. As we can see in Table \ref{tab:max}, for the full \textsc{GEAR} method using \textbf{bp1}, the results were around 1-2\% worse for all 3 ks in accuracy @ k. Similarly, for the other two prompts, we found a consistent under-performance when compared with averaging, again between 1\% and 2\% below, with the performance on Hill's test set going further below, up to 4\%. 

\begin{table}[ht]
\centering
\small
\begin{tabular}{llllcc}
\toprule
\textbf{Split} & \textbf{Prompt} & \textbf{ACC@1} & \textbf{ACC@10} & \textbf{ACC@100}\\
\midrule

\multirow{3}{*}{S.} &
bp1 & 63.6	& 81.2 & 94.0\\
& bp2 & 69.0 & 84.6	& 95.8\\
& rp  & 65.6	& 83.4 & 4.6\\ 
\cmidrule{1-5}
\multirow{3}{*}{U.} &
bp1 & 61.6	& 80.4 & 93.8 \\
& bp2 & 68.2 & 86.2 & 95.0 \\
& rp  & 66.6	& 85.2 & 95.2 \\
\cmidrule{1-5}
\multirow{3}{*}{D.} &
bp1 & 90.0 & 97.5 & 100\\
& bp2 & 90.5 & 97.0 & 100 \\
& rp  & 90.5 & 98.0 & 100\\

\bottomrule
\end{tabular}
\caption{Max pooling results across different prompts in Hill's dataset using the Instructor model for embeddings. S.: Seen split, U.: Unseen split, and D.: human description split. Prompt types are bp1 (Base Prompt 1), bp2 (Base Prompt 2), and rp (Reasoning Prompt).}
\label{tab:max}
\end{table}

\section{Conclusions and Future Work}

We have introduced a \textit{very} simple method for RD, which is based on a simple pipeline where we use LLMs to generate candidate terms given a definition, and embed them using some pooling technique (the best results are given by simple averaging). This unsupervised method outperforms existing supervised methods on a well known dataset, only falling short on the \textit{seen} split by methods that seem to overfit. In addition to this, we explored different components of this method, and evaluated its performance on various dictionaries and under different evaluation settings. For the future, we would like to explore more LLMs for the generation step, possibly developing a weighted average approach, which would be task-specific, and that could be learned via simple neural network architectures, similarly to \citet{wang2021deriving}, who used word classification datasets for tuning contextualized word embeddings. We would also like to expand our experiments to a multilingual setting, and test the outputs of an effective RD method as signal for pre-training general purpose embedding models. 

\section*{Limitations}

We have identified some limitations in this work. For instance, assuming that a single embedding model is effective across different registers and audiences might be an oversimplification. Considering various pre-training strategies could provide better insights into which dictionaries perform better or worse in different contexts. Additionally, although we have evaluated the performance of \textsc{GEAR} on the \texttt{3D-EX} dataset, comparing these results with baseline RD models is needed for a complete evaluation.

 

\bibliography{custom}

\appendix

\section{Prompt Types}
\label{sec:appendix:prompts}

\subsection{Base Prompt 1 (bp1):}

\begin{tcolorbox}
\footnotesize

Given the definition \{definition\}, generate a list of \{k\} terms defined by that definition assuming they are in \{dictionary\} dictionary. Only give me a list back, do not generate any other text.

\{dictionary\} is \{description\}

The returned list should follow the following conditions:
\begin{itemize}
    \item Terms should be ranked, with the first term being the most related to the definition.
    \item In a JSON object of the form \{ "terms": ["term\_1", "term\_2", \ldots ] \}.
    \item All terms should be in lowercase.
\end{itemize}

Example:

INPUT:
"A piece of furniture for sitting."

OUTPUT:
\{ "terms": ["chair", "stool", "bench", "sofa", "couch"] \}
\end{tcolorbox}

\subsection{Base Prompt 2 (bp2):}

\begin{tcolorbox}
\footnotesize

Given the definition \{definition\}, generate a list of \{k\} terms defined by that definition assuming they are in \{dictionary\} dictionary. Only give me a list back, do not generate any other text.

\{dictionary\} is \{description\}

These are some examples of definitions and terms in this dictionary: \{examples\}

The returned list should follow the following conditions:
\begin{itemize}
    \item Terms should be ranked, with the first term being the most related to the definition.
    \item In a JSON object of the form \{ "terms": ["term\_1", "term\_2", \ldots ] \}.
    \item All terms should be in lowercase.
\end{itemize}

Example:

INPUT:
"A piece of furniture for sitting."

OUTPUT:
\{ "terms": ["chair", "stool", "bench", "sofa", "couch"] \}
\end{tcolorbox}

\subsection{Reasoning Prompt (rp):}

\begin{tcolorbox}
\footnotesize

Given the definition \{definition\}, generate a list of \{k\} terms defined by that definition assuming they are in \{dictionary\} dictionary. Only give me a list back, do not generate any other text.

\{dictionary\} is \{description\}

These are some examples of definitions and terms in this dictionary: \{examples\}

For each term, provide an example usage in a sentence that matches the style and scope of \{dictionary\}.

The returned list should follow the following conditions:
\begin{itemize}
    \item Terms should be ranked, with the first term being the most related to the definition.
    \item All terms and examples should be in lowercase.
    \item Return the terms and examples in a JSON object of the form:
\end{itemize}

\{
  "terms": [
    \{ "term": "term\_1", "example": "example\_1" \},
    \{ "term": "term\_2", "example": "example\_2" \},
    ...
  ]
\}

Example:

INPUT:
"A piece of furniture for sitting."

OUTPUT:
\{
  "terms": [
    \{
      "term": "chair",
      "example": "he sat on the chair and opened his book."
    \},
    \{
      "term": "stool",
      "example": "she perched on the stool at the bar."
    \},
    \{
      "term": "bench",
      "example": "they rested on the bench after their walk."
    \},
    \{
      "term": "sofa",
      "example": "the family gathered on the sofa to watch TV."
    \},
    \{
      "term": "couch",
      "example": "he stretched out on the couch to take a nap."
    \}
  ]
\}
\end{tcolorbox}

\end{document}